\documentclass[10pt]{article}
\usepackage{float}
\usepackage{rotating}
\usepackage{changepage}
\usepackage[table]{xcolor}
\usepackage{nicefrac}
\usepackage[font=normalsize,format=plain,labelfont=bf,up,textfont=normal,up,justification=justified,singlelinecheck=false]{caption}
\usepackage{definition}
\usepackage{multirow}

\usepackage{verbatim}
\usepackage[toc,page]{appendix}
\usepackage{url}
\newcommand{\done}{\cellcolor{blue!15}}
\newcommand{\highlight}[1]{\colorbox{blue!15}{#1}}

\begin{document}
\title{Learning to Transfer Privileged Information}

\author{Viktoriia~Sharmanska$^1$\thanks{Corresponding Author: viktoriia.sharmanska@ist.ac.at},~Novi~Quadrianto$^2$, and~Christoph~Lampert$^1$,
		\vspace{0.5cm}\\
$^1$Institute of Science and Technology Austria, Austria\\
$^2$SMiLe CLiNiC, University of Sussex, UK}

\date{}

\maketitle

\begin{abstract}
\noindent
We introduce a learning framework called \emph{learning using privileged information (LUPI)} 
to the computer vision field. 
We focus on the prototypical computer vision problem of teaching computers to recognize objects in images. 
We want the computers to be able to learn faster at the expense of providing extra information during training time. 
As additional information about the image data, we look at several scenarios that have been studied in computer vision before: \emph{attributes}, \emph{bounding boxes} and \emph{image tags}.
The information is privileged as it is available at training time but not at test time. 
We explore two maximum-margin techniques that are able to make use of this additional source of information, for binary and multiclass object classification. 
We interpret these methods as learning easiness and hardness of the objects in the privileged space and then transferring this knowledge to train a better classifier in the original space. 
We provide a thorough analysis and comparison of information transfer from privileged to the original data spaces for both LUPI methods. 
Our experiments show that incorporating privileged information can improve the classification accuracy. 
Finally, we conduct user studies to understand which samples are easy and which are hard for human learning, and explore how this information is related to easy and hard samples when learning a classifier. 
%
%
%
%

%
\end{abstract}

\section{Introduction}
%
The framework called \emph{learning using privileged information (LUPI)} was introduced by V. Vapnik and A. Vashist in 2009 \cite{Vap09}, and it has not been recognized in the community until very recently. 
The concept is inspired by human experience of learning with teacher, when during learning we have access to training examples and to additional source of explanation from the teacher. 
For example learning a new concept in linear algebra is faster when the teacher explains it to us rather than if we get questions and right answers only. 
After the course, the students should be able to solve new tasks themselves and not rely on the teacher's expertise anymore. 
Training with teacher can significantly improve the learning process and ability to generalize in humans and machines \cite{Vap09}. 

Being introduces as a general framework, LUPI has been successfully applied to variety of tasks: 
image classification with different types of privileged data: attributes, bounding box annotation, textual description \cite{ShaQuaLam13}; 
handwritten digit images with poetic descriptions as privileged source for data clustering task \cite{Fey12}; facial feature detection with head pose or gender as privileged information \cite{Yan13}, facial expression recognition from low resolution images with high resolution images as privileged source \cite{Chen13}; metric learning \cite{Fou13}; counting with back-propagation \cite{Che14} etc. 

\begin{figure}[t]
\begin{center}
  \includegraphics[width=1.0\columnwidth]{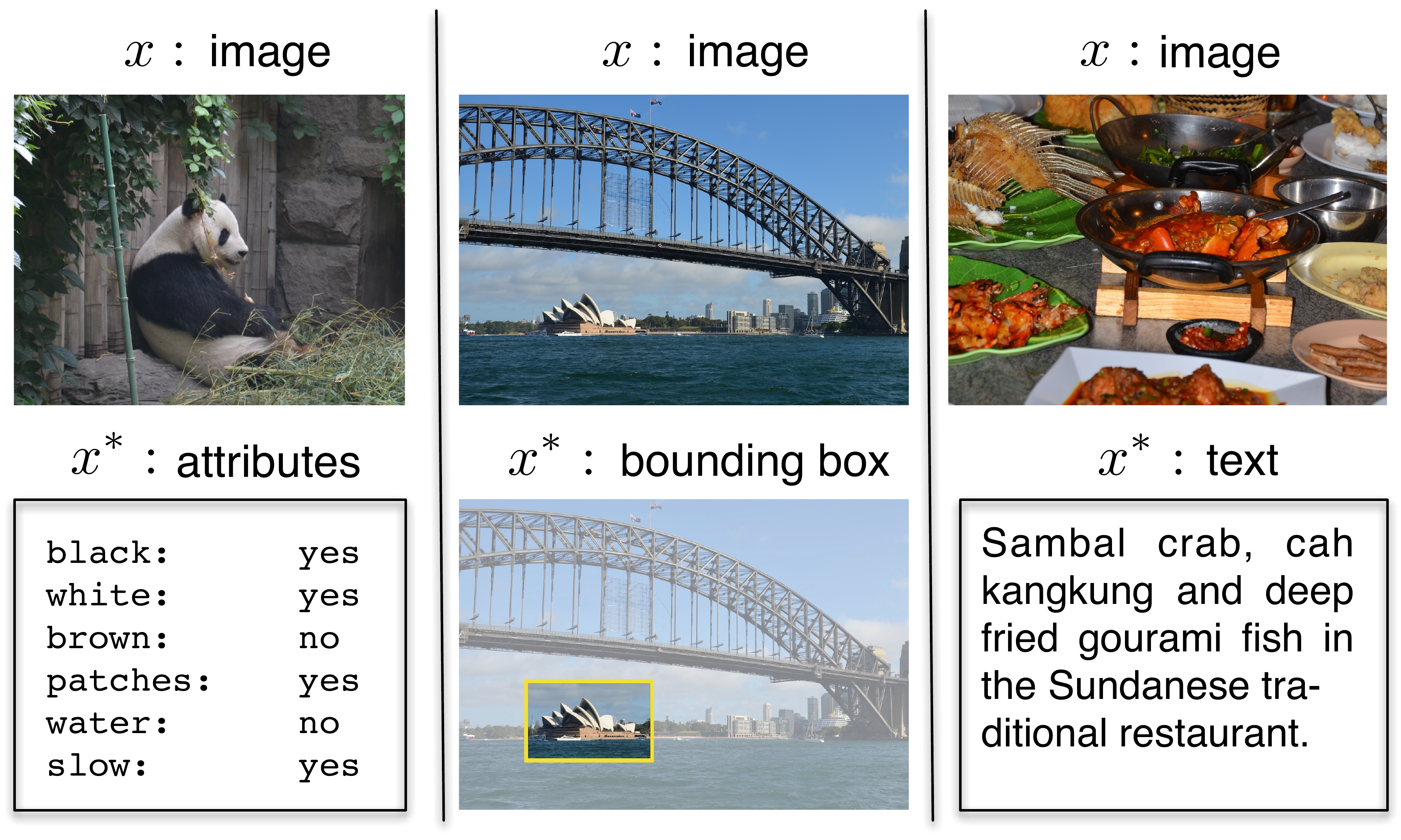}
\end{center}
   \caption{Three different forms of \emph{privileged} information 
   that can help learning better object recognition systems: 
   \emph{attributes}, \emph{object bounding boxes},
   and \emph{textual descriptions}.}
  \label{fig:motivation}
\end{figure}

In the standard learning setting, we are given input--output training pairs about the task we want to learn, 
for example, images and category labels for object classification. In the LUPI setting, we have the input--output training pairs plus additional information for each training pair that is \emph{only available during training}.
There is no direct limitation on the form of privileged information, 
i.e. it could be yet another feature representation, or completely different modality like text or hand annotation in addition to image data, that is specific for each training instance. 

LUPI in its original formulation does not tell us which kind of privileged information is useful, i.e. will lead to better performance, and how to measure the quality of it. 
In this work, which extends our original publication \cite{ShaQuaLam13}, we examine the three different types of privileged information in the context of object 
classification task: \emph{attributes} that describe semantic properties of an object, \emph{bounding boxes} that specify the exact localization of the target object in 
an image, and \emph{image tags} that describe the context of an image in textual form. 
Figure~\ref{fig:motivation} illustrates these three modalities.

\paragraph{Approach and contribution}
In order to do LUPI, we have to understand how to make use of the data modality that is not available at test time. 
For example, training a classifier on the privileged data is useless, since there 
is no way to evaluate the resulting classifier on the test data. 
%
At the core of our work lies the insight that \emph{privileged information 
allows us to distinguish between \emph{easy} and \emph{hard} examples 
in the training set.} 
Assuming that examples that are easy or hard with respect to the privileged information will also be easy or hard with 
respect to the original data, we enable \emph{information transfer} from the privileged to the original data modality. 
More specifically, we first define and identify which samples are easy and which are hard for the classification task, 
and incorporate the privileged information into the sample weights that encodes its easiness or hardness. 

We formalize the above observation in Section~\ref{sec:method}, where we study and compare two maximum-margin learning techniques for LUPI. 
The first, SVM+, was originally described by Vapnik~\cite{Vap09}. 
The second, \emph{Margin Transfer}, is our new contribution for the classification setting, and is an adaptation of the Rank Transfer method proposed in \cite{ShaQuaLam13} for ranking. 
We analyze the core difference of the information transfer in the proposed methods, 
and how this kind of knowledge about the learning problem can guide the training of an image-based predictor to a better solution.

In Section~\ref{sec:experiments}, we report on experiments in the three privileged information scenarios introduced earlier: attributes, bounding boxes and image tags. 
We demonstrate how to avoid handcrafted methods designed for specific type of additional information, and handle all the situations in a unified framework, which is another contribution of this work. 

In Section~\ref{sec:multiclass} we show that our method is naturally suitable for multiclass classification setting with one versus rest strategy.  
To our best knowledge, this is the first time LUPI methods were examined in the setting with more than two classes. 

We conduct user studies experiment to identify easy and hard samples in human learning of object categories. 
We then utilize this data as ground truth information to analyze and compare with easy and hard samples learned by the proposed LUPI methods in Section~\ref{sec:human}. 
This is the first attempt to understand when and which type of the privileged information could be useful. 

We end with the discussion and conclusions in Section~\ref{sec:conclusion}.
%
\section{Related work}\label{sec:relatedwork}
In computer vision problems it is common to have access to multiple 
sources of information. Sometimes all of them are visual, such as 
when images are represented by color features as well as by 
texture features. 
Sometimes, the modalities are mixed, such as for images with text captions.
If all modalities are present both at training and at test time, 
it is rather straight-forward to combine them for better prediction 
performance. This is studied, e.g., in the fields of \emph{multi-modal} 
or \emph{multi-view} learning. Methods suggested here range from \emph{stacking}, 
where one simply concatenates the feature vectors of all data modalities, 
to complex adaptive methods for early or late data fusions~\cite{snoek2005early}, 
including \emph{multiple kernel learning}~\cite{vedaldi2009multiple} and \emph{LP-$\beta$}~\cite{gehler2009feature}. 

Situations with an asymmetric distribution of information have also been explored. 
In \emph{weakly supervised} learning, the annotation available at training time 
is less detailed than the output one wants to predict. This situation occurs, 
e.g., when trying to learn an \emph{image segmentation} system using only 
per-image or bounding box annotation~\cite{kuettel2012segmentation}.
In \emph{multiple instance} learning, training labels are given not for 
individual examples, but collectively for groups of 
examples~\cite{maron1998multiple}. 
The inverse situation also occurs: for example in the PASCAL object recognition 
challenge, it has become a standard technique to incorporate strong annotation 
in the form of bounding boxes or per-pixel segmentations, even when the goal is 
just per-image object categorization~\cite{EveGooWilWinetal10,russakovsky2012object}. 
Similar to strong and weak supervision, situations in which the data representations 
differ between training and testing phase can be distinguished by whether one 
has less or more information available at training time than at test time.
The first situation occurs, e.g., in tracking, where temporal continuity can be 
used at test time that might not have been available at training time~\cite{kalal2009online}. 
Similarly, it has been shown that image metadata (geolocation, capture time)~\cite{chen2011clues} and an auxiliary feature modality~\cite{Sam14} can 
provide additional information at test time compared to only the image 
information available at training time. 

The situation we are interested in occurs when at training time we 
have an additional data representation compared to test time.
Different settings of this kind have appeared in the computer vision literature, 
but each was studied in a separate way. 
For example, for clustering with multiple image modalities, it has been proposed to use CCA 
to learn a shared representation that can be computed from either of representations~\cite{blaschko2008correlational}. 
Similarly the shared representation is also used for cross-modal retrieval~\cite{QuaLam11}.
Alternatively, one can use the training data to learn a mapping from the image to 
the privileged modality and use this predictor to fill in the values 
missing at test time~\cite{christoudias2008multi}.
Feature vectors made out of semantic attributes have been used to improve 
object categorization when very few or no training examples are available~\cite{Lam13,wang2010discriminative,sharmanska2012augmented,quadrianto2013supervised}.
%
In~\cite{Don11} it was shown that annotator rationales can act as additional sources of information 
during training, as long as the rationales can be expressed in the same 
data representation as the original data (e.g. characteristic regions 
within the training images). 

In \cite{Lob14}, the authors proposed to explore privileged information as measure of uncertainty about samples, 
estimating the noise term in the Gaussian Processes classification from the privileged data, i.e. privileged noise. 
%

Our work follows a different route than the above approaches. We are not looking
for task-specific solutions applicable to a specific form of privileged 
information.
Instead, we aim for a generic method that is applicable to any form of privileged  
information that is given as additional representations of the training data.

We show in the following sections that such frameworks do indeed exist, and 
in Section~\ref{sec:experiments} we illustrate that the individual 
situations described above can naturally be expressed in these frameworks.
%
\section{Learning using Privileged Information}
\label{sec:method}
In the following we will formalize the LUPI setup for the task of supervised binary classification. 
We describe a simple extension of LUPI for multiclass setting using one-versus-rest procedure in Section~\ref{sec:multiclass}.
Assume that we are given a set of $N$ training examples, represented by feature 
vectors $X=\{x_1,\dots,x_N\}\subset\Xcal=\RR^d$, their label annotation,
$Y=\{y_1,\dots,y_N\}\in\Ycal = \{ +1,-1\}$, and additional information, also in the form 
of feature vectors, $X^\ast=\{x^\ast_1,\dots,x^\ast_N\}\subset \Xcal^{*}=\RR^{d^\ast}$, 
where $x^\ast_i$ encodes the additional information we have 
about sample $x_i$. 
In the context of computer vision, we will consider the examples in $\Xcal$ as images 
and their features being extracted from the image content, for example, in a form of \emph{bag-of-visual-words} histograms~\cite{DanWilFanBraetal04}. 
We do not make any specific assumption about the \emph{privileged} data space $\Xcal^{*}$, 
and keep the general notation for the feature vectors extracted from visual, verbal or semantic form of privileged information. 
We will refer to $\Xcal$ and $\Xcal^{*}$ as original and privileged data spaces, accordingly. 

The binary classification task is to learn a prediction function $f:\Xcal\rightarrow\RR$ from a space 
$\Fcal$ of possible functions, e.g. all linear classifiers.  
The goal of LUPI is to use the privileged data, $X^\ast$, to 
learn a better classifier in the original data space $f:\Xcal\rightarrow\RR$, than one would learn without it. 
Since the privileged data is only available during training time and comes from a different domain, $\Xcal^\ast$, than the original space $\Xcal$, 
it is not possible, e.g., to apply functions defined on $\Xcal$ to $\Xcal^\ast$ or vice versa. 
In this work, we describe how to use the privileged data to characterize the training samples in the original data space into easy and hard cases. 
Knowing this will help us to direct the learning procedure towards better generalization and to learn a function of higher prediction quality. 

In the following, we explain two maximum-margin methods for learning 
with privileged information that fit to this interpretation. 
First method was proposed by Vapnik \emph{et al.}~\cite{Vap09} in $2009$, 
and second is our proposed alternative model for solving LUPI. 
For simplicity of notation we write all problems in their primal form. 
Kernelizing and dualizing them is possible using standard techniques~\cite{scholkopf2002learning}.

\subsection{Maximum Margin Model 1: SVM+}
The first model for learning with privileged information, SVM+, \cite{Vap09,Pec10} is based on a direct observation that a non-linearly separable (soft-margin) support vector machine (SVM) can be turned into a linearly separable (hard-margin) SVM if one had access to a so-called \emph{slack oracle}.
%
Standard soft-margin SVM classifier is trained based on the following constrained optimization problem:
\begin{subequations}
\label{eq:svm}
\begin{align}
\operatorname*{minimize}_{\substack{\xi_1,\dots,\xi_N\\\w \in\RR^{d}, b\in\RR}} 
&\dfrac{1}{2} \nbr{\w}^2 + C \sum_{i=1}^{N} \xi_i \label{eq:svm1}
\intertext{\hspace{3.15cm} subject to, for all $i = 1, \ldots, N$,}
&\hspace{-1.5cm} y_i [ \left\langle \w, \x_i \right\rangle + b] \geq 1 - \xi_i \quad \text{and}\quad  \xi_i\geq 0. \label{eq:svm2}
\end{align}
\end{subequations}
We note that our SVM classifier is fully characterized by its 
weight vector $\w$ and bias parameter $b$. 
However, at training phase, $N$ slack variables $\xi_i$ -- one for each training sample -- also need to be estimated.
When the number of training examples increases, soft-margin SVM solutions are known 
to converge with a rate of $\BigO{\frac{1}{\sqrt{N}}}$ to the optimal classifier~\cite{vapnik1999nature}.
This is in sharp contrast to hard-margin solutions that converge with a rate of $\BigO{\frac{1}{N}}$.
One then wonders whether it is possible for a soft-margin SVM to have a faster convergence rate, ideally at
the same rate as hard-margin SVM.
If the answer is positive, the improved soft-margin SVM would require fewer training examples to reach 
a certain prediction accuracy than a standard one. 
Intuitively, with $\BigO{\frac{1}{N}}$ rate, we will only
require $100$ samples instead of $10,000$ to achieve the same level of predictive performance.

It might not come as a surprise that if we knew the optimal slack values $\xi_i$ in 
the optimization problem \eq{eq:svm}, for example from an \emph{oracle}, 
then the formulation can be reduced to the hard-margin case with the convergence rate $\BigO{\frac{1}{N}}$, 
when fewer parameter need to be inferred from the data~\cite{Vap09}. 
Instead of $N+d+1$ unknowns which include slack variables, we are now estimating only 
$d+1$ unknowns which are the actual object of interest, our classifying hyperplane. 
The interpretation of slack variables is to tell us which training examples 
are \emph{easy} and which are \emph{hard} and in the above OracleSVM, we do not have 
to infer those variables from the data as they are given by the oracle.

The idea of the SVM+ classifier is to use the privileged information as a proxy to the oracle. 
For this we parameterize the slack for $i$-th sample $\xi_i=\langle \w^{*}, \x_i^{*} \rangle + b^{*}$ with unknown $\w^\ast$ 
and $b^\ast$, obtaining the SVM+ training problem:
\begin{subequations}
\label{eq:svmplus}
\begin{align}
\operatorname*{minimize}_{\substack{\w \in\RR^{d}, b\in\RR \\ \w^\ast \in\RR^{d^\ast}, b^\ast \in\RR}}  
&\frac{1}{2}\left( \nbr{\w}^2 +\!\gamma\!\nbr{ \w^\ast}^2 \right) 
+ C \sum_{i=1}^{N}  \langle \w^\ast, \x_i^\ast \rangle + b^\ast \label{eq:svmplus1}
\intertext{\hspace{2.25cm} subject to, for all $i = 1, \ldots, N$,}
&y_i [\langle \w, \x_i \rangle + b] \geq 1 - [ \langle \w^\ast, \x_i^\ast \rangle + b^\ast ] \label{eq:svmplus2}
\\
\text{ and }\qquad&\langle \w^\ast, \x_i^\ast \rangle + b^\ast \geq 0. \label{eq:svmplus3}
\end{align}
\end{subequations}
The above SVM+ parameterizes the slack variables with a finite hypothesis space (a scalar and a weight vector with dimension $d^\ast$, for example), 
instead of allowing them to grow linearly with the number of examples $N$. 

\paragraph{Numerical optimization}

The SVM+ optimization problem \eq{eq:svmplus} is convex, 
and can be solved in the dual representation using a standard quadratic programming (QP) solver.
For a medium size problem (thousands to hundreds of thousands), a general purpose QP solver might not 
suffice, and special purpose algorithms have to be developed to solve the QP. 
In~\cite{Pech11}, suitable sequential minimal optimization (SMO) 
algorithms were derived to tackle the problem. 
However, for the problem size that we are experimenting with (hundreds of samples), 
we find that using a general purpose QP provided in the \texttt{CVXOPT}\footnote{\url{http://cvxopt.org}} package is faster than 
the specialized SMO solver. Therefore, we use the \texttt{CVXOPT}-based QP solver for our experiments (Section~\ref{sec:experiments}). 
%

\subsection{Maximum Margin Model 2: Margin Transfer}

In this work, we propose a second model called \emph{Margin Transfer} that: 1) can be solved by a sequence of standard SVM solvers; and 2) \emph{explicitly} enforces an easy-hard interpretation for transferring information from the privileged to the original space.  
The method is an adaptation of the \emph{Rank Transfer} method previously described in \cite{ShaQuaLam13} for the ranking setup, where we identify and transfer the information about easy-to-separate and hard-to-separate pairs of examples. 
Here, we propose to follow a similar strategy but instead of looking at pairs of examples, we check each example whether it is easy-to-classify or hard-to-classify based on the margin distance to the classifying hyperplane in the privileged space. Subsequently, we transfer this knowledge to the original space. 
We hypothesize that knowing a priori which examples are easy to classify and which are hard during 
learning should improve the prediction performance. 
This consideration leads us to the Margin Transfer method, summarized in Algorithm~\ref{alg:MarginTrans}.

First, we train an ordinary SVM on $X^\ast$. 
The resulting prediction function $f^\ast(\x^\ast) = \inner{\w^\ast}{\x^\ast}$ 
is used to compute the margin distance from the training samples to the classifying hyperplane in the privileged space\footnote{Note that in the standard SVM formulation one would compute the values of slack variables to know how far is the sample from the hyperplane. 
As slack variables appear only at training phase, we deliberately evaluate the prediction function on the same data it was trained on to identify easy and hard samples \emph{at train} time.} $\rho_{i}:= y_i f^\ast(\x^\ast_i)$. 
%
Examples with a large values of $\rho_{i}$ are considered easy to classify, 
whereas small or even negative values of $\rho_{i}$ indicate hard or even impossible 
to classify samples. 
We then train a standard SVM on $X$, aiming for a \emph{data-dependent margin} $\rho_{i}$ 
transferred from the privileged space rather than enforcing a constant margin of $1$. 
The corresponding optimization problem is
\begin{subequations}
\label{eq:margintransfer}
\begin{align}
&\operatorname*{minimize}_{\w \in\RR^{d},\: \xi_{i} \in\RR}\quad \dfrac{1}{2} \nbr{\w}^2 + C \sum_{i=1}^{N} \xi_{i} \label{eq:margintransfer1}
\intertext{\hspace{3.5cm} subject to, for all $i = 1, \ldots, N$ }
&y_i\left\langle \w, \x_{i}\right\rangle  \geq \rho_{i} - \xi_{i} 
\quad\text{and}\quad\xi_{i}\geq 0.\label{eq:margintransfer2}
\end{align}
\end{subequations}
One can see that examples with small and negative values of $\rho_{i}$ have limited 
influence on $\w$ comparing to the standard SVM, because their slacks $\xi_{i}$ can easily compensate for the inequality constraint. 
We threshold the negative values of margin at certain tolerance value $\epsilon$, $\epsilon \geq 0$. 
Our interpretation is that if it was not possible to correctly classify a sample in the privileged space, 
it will also be not possible to do so in the, presumably weaker, 
original space. 
Forcing the optimization to solve a hopeless tasks would only lead 
to overfitting and reduced prediction accuracy. 
\begin{algorithm}[t]
\caption{Margin Transfer from $\Xcal^\ast$ to $\Xcal$}\label{alg:MarginTrans}
\begin{algorithmic}
\STATE {\bfseries Input} original data $X$, privileged data $X^*$, labels $Y$, tolerance $\epsilon \geq 0$ 
\STATE $f^\ast\leftarrow$ SVM \eq{eq:svm} trained on $(X^{*},Y)$
\STATE $\rho_{i}= \text{max} \:\{y_i f^\ast(\x^\ast_i),\: \epsilon\}$ \quad \textit{(per-sample margin)}
\STATE $f\leftarrow$ SVM \eq{eq:margintransfer} trained on $(X,Y)$ using $\rho_{i}$ instead of unit margin.
\STATE {\bfseries Return} $f:\Xcal\to\RR$
\end{algorithmic}
\end{algorithm}

\paragraph{Numeric Optimization}
Both learning steps in the \emph{Margin Transfer} method, are convex optimization problems. 
Furthermore, in contrast to SVM+, we can use standard SVM packages
to solve them, including efficient methods working in primal 
representation~\cite{chapelle2007training}, and solvers based 
on stochastic gradient descent~\cite{shalev2007pegasos}. 

For the SVM with data-dependent margin \eq{eq:margintransfer1}-\eq{eq:margintransfer2}, we do the following reparameterization: 
we divide each constraint \eq{eq:margintransfer2} by the corresponding $\rho_{i}$, which is possible after thresholding at the non-negative tolerance value. 
For our experiments, we threshold at $\epsilon=0.1$, thereby preventing numeric instabilities and increasing computational efficiency of the method. 
Changing variables from $\x_{i}$ to $\hat \x_{i}=\frac{\x_{i}}{\rho_{i}}$
and from $\xi_{i}$ to $\hat\xi_{i}=\frac{\xi_{i}}{\rho_{i}}$ we 
obtain the equivalent optimization problem
\begin{subequations}
\begin{align}
&\operatorname*{minimize}_{\w \in\RR^{d},\: \hat{\xi}_{i} \in\RR}\quad \dfrac{1}{2} \nbr{\w}^2 + 
C \sum_{i=1}^{N} \rho_{i} \hat{\xi}_{i} 
\intertext{\hspace{3.5cm} subject to, for all $i = 1, \ldots, N$}
&y_i\inner{\w }{\hat \x_{i}}  \geq 1 - \hat{\xi}_{i} 
\quad \text{and} \quad \hat{\xi}_{i}\geq 0.
\end{align}
\end{subequations}
This corresponds to standard SVM optimization with training examples 
$\hat \x_{i}$, where each slack variable has 
an individual weight $C\rho_{i}$ in the objective. 
Many existing SVM packages support such per-sample weights, in 
our experiments we use \emph{\sc liblinear}~\cite{REF08a}. 
Additionally we would like to position our model in support of recent results that SVM+ classifiers can be reformulated as a special forms of example-weighted binary SVMs~\cite{Lap14}.
%
%
\subsection{How is Information being Transferred?}
We elaborate on how SVM+ and Margin Transfer instantiate the easy-hard interpretation and how they differ from each other. 

\textbf{Observation 1: Both methods, SVM+ and Margin Transfer, concentrate on learning easy samples and deemphasizing the hard ones.} 

Though SVM+ and Margin Transfer aim at the same goal, the way this is achieved is different in these two methods. 
Let us illustrate this by using the oracle analogy. 
In the SVM+, the oracle gives us the value of the slack function $\texttt{oracle}_{\text{svm+}} (\x_i) := \langle \w^{*}, \x_i^{*} \rangle + b^{*}$ for example $\x_i$, 
and in the Margin Transfer, the oracle gives us the margin distance to the classifying hyperplane $\texttt{oracle}_{\text{margin transfer}} (\x_i) := y_i f^\ast(\x^\ast_i)$. 

Suppose we only have two training samples, $\x_1$ and $\x_2$, and 
we ask the oracles what they perceive about the two samples. 
Say, in case of SVM+, we get back the following answers: $\texttt{oracle}_{\text{svm+}} (\x_1) = 10.0$ and $\texttt{oracle}_{\text{svm+}} (\x_2) = 0.0$. 
This means that the first sample is hard (its slack variable is high) and the second one is easy (its slack variable is zero). 
When we encode this into the optimization problem of SVM+, we can see that the constraint \eq{eq:svmplus2} becomes 
$y_1 [\inner{\w}{\x_1} + b] \geq -9$, (effort\emph{less} to satisfy comparing to the unit margin in the standard SVM) for the first sample and  $y_2 [\inner{ \w}{\x_2} + b] \geq 1$ (effort\emph{ful} to satisfy comparing to the standard SVM) for the second one. 
So this means that the optimization task would more or less ignore the constraint of the first sample (that is hard) and concentrate to satisfy the constraint about the second sample (that is easy).

We repeat the questions to the Margin Transfer oracle and say the answers are: \\
$\texttt{oracle}_{\text{margin transfer}} (\x_1) = -5$ and $\texttt{oracle}_{\text{margin transfer}} (\x_2) = 8$. 
Interpreting oracle's answers lead us to conclude that the first sample is hard (its margin distance is zero or negative) and the second one is easy (its margin distance is positive). 
When we encode this into the optimization problem of Margin Transfer, the constraint \eq{eq:margintransfer2} becomes 
$y_1\inner{ \w}{ \x_{1}}  \geq \epsilon - \xi_{1}$ (effortless to satisfy)  for the first sample and  $y_2\inner{ \w}{ \x_{2}}  \geq 8 - \xi_{2}$ (effortful to satisfy) for the second one. 
The same as before, the optimization task would ignore the constraints of the hard samples and concentrate on learning the easy ones.
This is despite the fact that SVM+ oracle returns high value for hard samples while margin transfer oracle returns low value for hard samples, and vice versa for easy ones.

\textbf{Observation 2: Classification performance in the privileged space matters for Margin Transfer but not for SVM+.}

At the core of SVM+ lies the idea of imitating the oracle by learning the non-negative linear regression slack function defined in the privileged space. 
The information about labels does not come into play when modeling the slack function, 
so in a sense, we never validate the classification performance in the privileged space.  
In contrast to it, in the Margin Transfer method, 
the performance in the privileged space explicitly guides the training of the predictor in the original data space. 
Samples that are easy and hard to classify in the privileged space directly define the margin for the samples in the original data space. 

In the Rank Transfer method described in \cite{ShaQuaLam13} we observe another way to do information transfer considering pairs of samples. 
For any pair of samples from different classes we estimate whether it is easy-to-separate or hard-to-separate pair based on the rank margin between the samples in the privileged space. 
We transfer this information into the ranking SVM objective and completely ignore pairs that got swapped. 
In the rank transfer framework, we deal with pairs of samples and therefore suffer from quadratic amount of constraints to be satisfied if every pair of samples to be considered. 


\section{Experiments}
\label{sec:experiments}
In our experimental setting we study three different types of privileged information, showing that all of these can be handled in a unified framework, where previously hand crafted methods were used. 
We consider attribute annotation, bounding box annotation and textual description as sources of privileged information if these are present at training time but not at test time. 
As we will see, some modalities are more suitable for transferring the margin than others. 
We will discuss this in the following subsections. 

\textbf{Methods.}
We analyze two methods of learning using privileged information: our proposed Margin Transfer method for transferring the margin, 
and the SVM+ method \cite{Pech11}. 
We compare the results with ordinary SVM when learning on the original space $\Xcal$ directly. 
We also provide as a reference the performance of SVM in the privileged space $\Xcal^{*}$, 
as if we had the access to the privileged information during testing. 

\textbf{Evaluation metric.}
To evaluate the performance of the methods we use accuracy, and we report mean and standard error across $20$ repeats. 

\textbf{Model selection.}
For the LUPI methods, 
we perform a joint cross validation model selection approach for choosing the regularization parameters in the original and privileged spaces. 
In the SVM+ method these are $C$ and $\gamma$ \eq{eq:svmplus1}, and in the Margin Transfer these are $C$'s in the two-stage procedure \eq{eq:svm1}, \eq{eq:margintransfer1}. 
For the methods that do not use privileged information there is only a regularization parameter $C$ to be cross validated. 
In the privileged space we select over $7$ parameters $\{10^{-3}, \dots, 10^{3}\}$. 
We use the same range in the original space if the data is $L_2$ normalized, and the range $\{10^{0}, \dots, 10^{5}\}$ for $L_1$ normalized data. 
In our experiments we use 5x5 fold cross-validation scheme for binary classification and 5 fold cross-validation for multiclass setting. 
The best parameter (or pair of parameters) found is used to retrain the complete training set. 
Based on our experience, LUPI methods require very thorough model selection step. 
To couple the modalities of privileged and original data spaces properly, the grid search over both parameter spaces has to be exploited. 
\subsection{Attributes as privileged information} 
\label{sec:attributes}
Attribute annotation incorporates high-level description of the semantic properties of different objects like shape, color, habitation forms etc. 
The concept of attributes was introduced in \cite{Lam09}, \cite{Far09} for attribute-based classification, when object in the image is classified based on attributes it has. 
We use the \emph{Animals with Attributes (AwA)}\footnote{\url{http://attributes.kyb.tuebingen.mpg.de/}} dataset~\cite{Lam09,Lam13}. 
We focus on the default $10$ test classes, for which the attribute annotation is provided together with the dataset.
The $10$ classes are \emph{chimpanzee, giant panda, leopard, persian cat, pig, hippopotamus, humpback whale, raccoon, rat, seal}, and contain $6180$ images in total. 
The attributes capture $85$ properties of the animals, color, texture, shape, body parts, behavior among others. 
We use $L_1$ normalized $2000$ dimensional SURF descriptors \cite{Bay08} as original features, and $85$ dimensional predicted attributes as the privileged information. 
The values of the predicted attributes are obtained from the DAP model \cite{Lam13} and correspond to probability estimates of the binary attributes in the images. 
We train $45$ binary classifiers, one for each pair of the $10$ classes with $100$ images per class as training data. 
We use $200$ samples per class for testing. 
To get better statistics of the performance we repeat the procedure of train/test split $20$ times. 

\begin{figure*}[th]
\begin{center}
  \includegraphics[width=0.95\columnwidth,page=1]{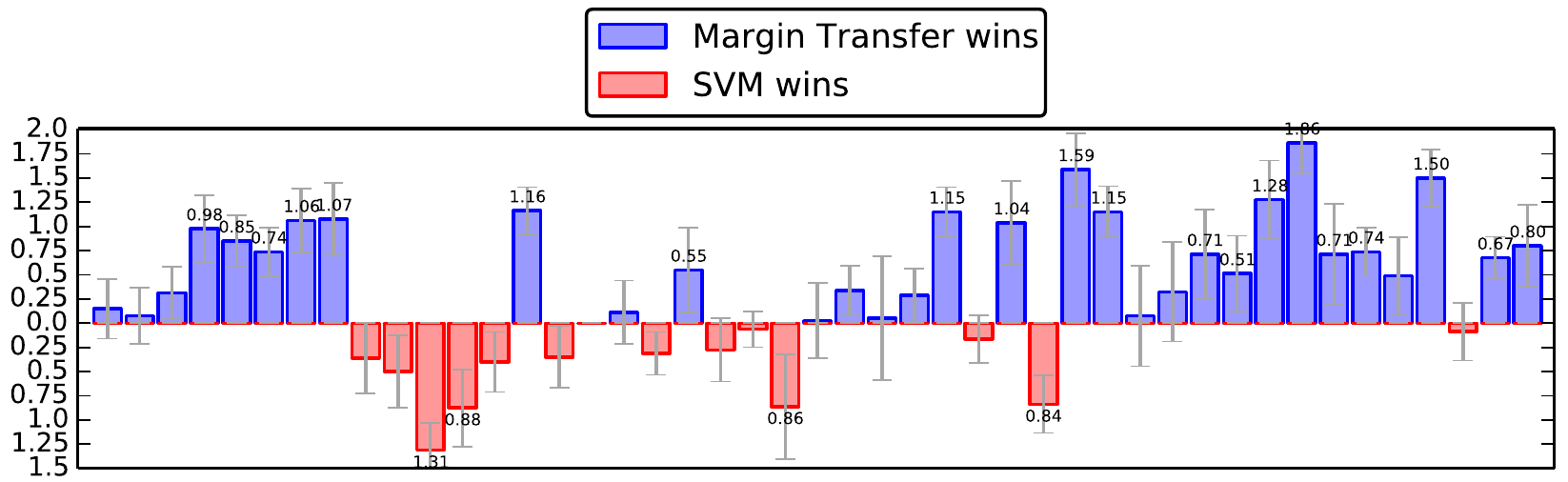}
  \includegraphics[width=0.95\columnwidth,page=2]{AwA_full.pdf}
\end{center}
   \caption{AwA dataset (attributes as privileged information). 
Pairwise comparison of the methods that utilize privileged information (Margin Transfer, SVM+) and their baseline counterpart (SVM) is shown via difference of the accuracy performance. 
The length of the $45$ bars corresponds to relative improvement of the accuracy for $45$ cases.} 
\label{fig:AwA}
\end{figure*}

\begin{table*}[thp]	
\centering
\scalebox{0.7}{
\setlength{\tabcolsep}{4pt}
\begin{tabular}{c|l|c|c|c|c|c|c}		          				          
&					& {\tt SVM }  		& {\tt Margin Transfer} & {\tt SVM+}  		&& {\tt Reference}   	\\
&					&  image 		&  image+attributes	& image+attributes   	&& \it{(SVM attributes)} 	\\\hline
1 & Chimpanzee versus Giant panda 	& $83.25 \pm 0.53$ 	& $83.40 \pm 0.43$ 	& $\bf{83.77 \pm 0.48}$ && $\it{85.00 \pm 0.42}$\\
2 & Chimpanzee versus Leopard 		& $86.63 \pm 0.35$ 	& $86.71 \pm 0.38$ 	& $\bf{86.76 \pm 0.35}$ && $\it{92.95 \pm 0.27}$\\
3 & Chimpanzee versus Persian cat 	& $83.91 \pm 0.46$ 	& $\bf{84.22 \pm 0.41}$ & $83.93 \pm 0.49$ 	&& $\it{91.42 \pm 0.31}$\\
4 & Chimpanzee versus Pig 		& $79.72 \pm 0.35$ 	& $\done\bf{80.70 \pm 0.26}$ 	& $80.55 \pm 0.27$ 	&& $\it{86.53 \pm 0.43}$\\
5 & Chimpanzee versus Hippopotamus 	& $81.05 \pm 0.28$ 	& $\done\bf{81.90 \pm 0.27}$	& $81.78 \pm 0.29$ 	&& $\it{88.12 \pm 0.29}$\\
6 & Chimpanzee versus Humpback whale 	& $94.45 \pm 0.26$ 	& $\done\bf{95.18 \pm 0.21}$	& $94.75 \pm 0.24$ 	&& $\it{98.32 \pm 0.16}$\\
7 & Chimpanzee versus Raccoon 		& $80.11 \pm 0.50$ 	& $\done\bf{81.17 \pm 0.48}$	& $80.68 \pm 0.45$ 	&& $\it{85.47 \pm 0.38}$\\
8 & Chimpanzee versus Rat 		& $80.15 \pm 0.43$ 	& $\done\bf{81.22 \pm 0.43}$	& $81.21 \pm 0.42$ 	&& $\it{90.03 \pm 0.48}$\\
9 & Chimpanzee versus Seal 		& $\bf{85.80 \pm 0.26}$ & $85.43 \pm 0.44$ 	& $85.65 \pm 0.31$ 	&& $\it{91.12 \pm 0.24}$\\
10 & Giant panda versus Leopard 	& $\bf{87.82 \pm 0.32}$ & $87.32 \pm 0.37$ 	& $87.10 \pm 0.35$ 	&& $\it{92.52 \pm 0.31}$\\
11 & Giant panda versus Persian cat 	& $\bf{87.66 \pm 0.37}$ & $86.35 \pm 0.30$ 	& $87.10 \pm 0.28$ 	&& $\it{88.92 \pm 0.40}$\\
12 & Giant panda versus Pig 		& $\bf{80.80 \pm 0.44}$ & $79.92 \pm 0.41$ 	& $80.25 \pm 0.39$ 	&& $\it{84.57 \pm 0.40}$\\
13 & Giant panda versus Hippopotamus 	& $\bf{85.36 \pm 0.41}$ & $84.96 \pm 0.51$ 	& $84.32 \pm 0.43$ 	&& $\it{90.36 \pm 0.34}$\\
14 & Giant panda versus Humpback whale 	& $94.30 \pm 0.30$ 	& $\done\bf{95.46 \pm 0.25}$ & $95.36 \pm 0.24$ 	&& $\it{98.36 \pm 0.15}$\\
15 & Giant panda versus Raccoon 	& $\bf{83.52 \pm 0.44}$ & $83.17 \pm 0.44$ 	& $83.18 \pm 0.44$ 	&& $\it{84.08 \pm 0.30}$\\
16 & Giant panda versus Rat 		& $81.76 \pm 0.45$ 	& $81.76 \pm 0.38$ 	& $\bf{81.96 \pm 0.43}$ && $\it{87.60 \pm 0.27}$\\
17 & Giant panda versus Seal 		& $85.47 \pm 0.41$ 	& $85.58 \pm 0.40$ 	& $\bf{85.98 \pm 0.31}$ && $\it{89.42 \pm 0.35}$\\
18 & Leopard versus Persian cat 	& $\bf{90.18 \pm 0.23}$ & $89.87 \pm 0.26$ 	& $89.71 \pm 0.28$ 	&& $\it{93.32 \pm 0.25}$\\
19 & Leopard versus Pig 		& $81.20 \pm 0.42$ 	& $\bf{81.75 \pm 0.43}$ & $80.71 \pm 0.22$ 	&& $\it{91.01 \pm 0.26}$\\
20 & Leopard versus Hippopotamus 	& $\bf{86.37 \pm 0.33}$ & $86.10 \pm 0.31$ 	& $86.05 \pm 0.26$ 	&& $\it{91.36 \pm 0.31}$\\
21 & Leopard versus Humpback whale 	& $95.26 \pm 0.34$ 	& $95.20 \pm 0.36$ 	& $\bf{95.77 \pm 0.20}$ && $\it{98.60 \pm 0.10}$\\
22 & Leopard versus Raccoon 		& $77.40 \pm 0.51$ 	& $76.53 \pm 0.76$ 	& $\bf{77.46 \pm 0.50}$ && $\it{81.36 \pm 0.35}$\\
23 & Leopard versus Rat 		& $81.82 \pm 0.26$ 	& $\bf{81.85 \pm 0.40}$ & $81.33 \pm 0.35$ 	&& $\it{90.55 \pm 0.23}$\\
24 & Leopard versus Seal 		& $87.28 \pm 0.36$ 	& $87.62 \pm 0.36$ 	& $\bf{87.67 \pm 0.31}$ && $\it{92.36 \pm 0.31}$\\
25 & Persian cat versus Pig 		& $76.28 \pm 0.61$ 	& $\bf{76.33 \pm 0.62}$ & $76.26 \pm 0.58$ 	&& $\it{78.16 \pm 0.35}$\\
26 & Persian cat versus Hippopotamus 	& $84.85 \pm 0.56$ 	& $\bf{85.13 \pm 0.46}$ & $84.27 \pm 0.48$ 	&& $\it{90.62 \pm 0.35}$\\
27 & Persian cat versus Humpback whale 	& $91.81 \pm 0.32$ 	& $\done\bf{92.96 \pm 0.26}$ & $92.66 \pm 0.30$ 	&& $\it{97.87 \pm 0.16}$\\
28 & Persian cat versus Raccoon 	& $84.50 \pm 0.44$ 	& $84.33 \pm 0.44$ 	& $\bf{84.76 \pm 0.42}$ && $\it{85.65 \pm 0.32}$\\
29 & Persian cat versus Rat 		& $65.32 \pm 0.38$ 	& $\done\bf{66.36 \pm 0.51}$ & $65.87 \pm 0.48$ 	&& $\it{65.51 \pm 0.71}$\\
30 & Persian cat versus Seal 		& $80.61 \pm 0.38$ 	& $79.77 \pm 0.48$ 	& $\bf{80.76 \pm 0.57}$ && $\it{86.65 \pm 0.35}$\\
31 & Pig versus Hippopotamus 		& $71.26 \pm 0.38$ 	& $\done\bf{72.85 \pm 0.42}$ & $71.70 \pm 0.37$ 	&& $\it{77.77 \pm 0.61}$\\
32 & Pig versus Humpback whale 		& $91.31 \pm 0.38$ 	& $\done\bf{92.46 \pm 0.32}$ & $91.95 \pm 0.44$ 	&& $\it{97.41 \pm 0.14}$\\
33 & Pig versus Raccoon 		& $75.63 \pm 0.51$ 	& $\bf{75.71 \pm 0.39}$ & $75.03 \pm 0.28$ 	&& $\it{82.28 \pm 0.36}$\\
34 & Pig versus Rat 			& $67.80 \pm 0.44$ 	& $\bf{68.12 \pm 0.40}$ & $67.37 \pm 0.59$ 	&& $\it{74.12 \pm 0.38}$\\
35 & Pig versus Seal 			& $76.10 \pm 0.50$ 	& $\bf{76.81 \pm 0.38}$ & $75.83 \pm 0.37$ 	&& $\it{82.11 \pm 0.24}$\\
36 & Hippopotamus versus Humpback whale & $86.67 \pm 0.40$ 	& $\bf{87.18 \pm 0.38}$ & $86.38 \pm 0.38$ 	&& $\it{95.02 \pm 0.25}$\\
37 & Hippopotamus versus Raccoon 	& $80.61 \pm 0.61$ 	& $\done\bf{81.88 \pm 0.60}$ & $80.92 \pm 0.52$ 	&& $\it{87.62 \pm 0.28}$\\
38 & Hippopotamus versus Rat 		& $77.83 \pm 0.45$ 	& $\done\bf{79.70 \pm 0.41}$ & $78.25 \pm 0.48$ 	&& $\it{88.41 \pm 0.37}$\\
39 & Hippopotamus versus Seal 		& $67.91 \pm 0.60$ 	& $\bf{68.62 \pm 0.55}$ & $67.98 \pm 0.49$ 	&& $\it{73.32 \pm 0.50}$\\
40 & Humpback whale versus Raccoon 	& $92.52 \pm 0.22$ 	& $\done\bf{93.26 \pm 0.30}$ & $92.66 \pm 0.30$ 	&& $\it{97.21 \pm 0.19}$\\
41 & Humpback whale versus Rat 		& $89.22 \pm 0.45$ 	& $\bf{89.71 \pm 0.33}$ & $89.27 \pm 0.37$ 	&& $\it{97.27 \pm 0.15}$\\
42 & Humpback whale versus Seal 	& $80.67 \pm 0.37$ 	& $\done\bf{82.17 \pm 0.34}$ & $81.15 \pm 0.33$ 	&& $\it{89.37 \pm 0.34}$\\
43 & Raccoon versus Rat 		& $\bf{73.75 \pm 0.38}$ & $73.66 \pm 0.43$ 	& $73.22 \pm 0.52$ 	&& $\it{77.92 \pm 0.36}$\\
44 & Raccoon versus Seal 		& $83.86 \pm 0.42$ 	& $\done\bf{84.53 \pm 0.41}$ & $84.02 \pm 0.45$ 	&& $\it{87.32 \pm 0.25}$\\
45 & Rat versus Seal 			& $74.03 \pm 0.56$ 	& $\bf{74.83 \pm 0.52}$ & $73.93 \pm 0.46$ 	&& $\it{86.46 \pm 0.23}$
\end{tabular}	}
\parbox{14cm}{
\caption{
AwA dataset (attributes as privileged information). 
The numbers are mean and standard error of accuracy over $20$ runs. 
The best result is highlighted in \textbf{boldface}, which in total is $\bf{9}$ for {\tt SVM}, $\bf{27}$ for {\tt Margin Transfer}, and $\bf{9}$ for {\tt SVM+}. 
Highlighted \highlight{blue} indicates significant improvement of the methods that utilize privileged information (Margin Transfer and/or SVM+) over the methods that do not (SVM). 
We used a paired Wilcoxon test with $95\%$ confidence level as a reference. 
Additionally, we also provide the SVM performance on $\Xcal^{*}$ (last column).
\label{tab:AwA}}}
\end{table*}

\textbf{Results.} 
As we can see from the Figure~\ref{fig:AwA}, utilizing attributes as privileged information for object classification task is useful. 
Margin Transfer outperforms SVM in $32$ out of $45$ cases, and SVM+ outperforms SVM in $27$ out of $45$ cases. 
Noticeably, the Margin Transfer model is able to utilize privileged information better than the SVM+. 
We observe partial overlap of cases where Margin Transfer and SVM+ are not able to utilize privileged information (location of the red bars). 
The red bars coincide mostly in pairs with Giant panda, Leopard versus other animals.
Full comparison of the accuracy of all methods is shown in the Table \ref{tab:AwA}. 
We also notice, that the gain of the Margin Transfer method is higher in the regime when the problem is hard, i.e. when accuracy is below $90\%$. 
As a further analysis, we also check the hypothetical performance of SVM in the privileged space $\Xcal^{*}$. 
The privileged information has consistently higher accuracy than SVM in the original space $\Xcal$. 
In most cases, higher accuracy in the privileged space than in the original space translates to positive effect in margin transfer. 
We credit this to the fact, that Margin Transfer relies on the performance in the privileged space in order to explore easiness and hardness of the samples. 
And it is successful if the underlying assumption that the same examples are easy and hard in both modalities is fulfilled, 
as it is in most of the cases here. 
%

\subsection{Bounding box as privileged information}
\label{sec:bbox}
Bounding box annotation is designed to capture the exact location of an object in the image. It is usually represented as a box around the object. 
When performing image-level object recognition, knowing the exact location of the object in the training data is privileged information. 
We use a subset of the categories from the ImageNet 2012 challenge (ILSVRC2012) for which bounding box annotation 
is available\footnote{\url{http://www.image-net.org/challenges/LSVRC/2012/index}}. 
We define two groups of interest:  group with variety of snakes, and group with balls in different sport activities. 
The group of snakes has $17$ classes: \emph{thunder snake, ringneck snake, hognose snake, green snake, king snake, garter snake, water snake, vine snake, night snake, boa constrictor, rock python, indian cobra, green mamba, sea snake, horned viper, diamondback, sidewinder}, and has $8254$ images in total, on average $500$ samples per class. 
We ignore few images with too small bounding box region, 
and use $8227$ images for further analysis. 
The group balls has $6$ classes: \emph{soccer ball, croquet ball, golf ball, ping-pong ball, rugby ball, tennis ball}, and has $3259$ images in total, on average $500$ samples per class. Here, we also ignore images with uninformative bounding box annotation and use $3165$ images instead. 
We consider one-versus-rest scenario for each group separately.
We use $L_2$ normalized $4096$-dimensional Fisher vectors~\cite{PerSanMen10} extracted from the whole 
images as well as from only the bounding box regions, and we use the former 
as the original data representation and the latter as privileged 
information. 
We train one binary classifier for each class, $17$ in the first group and $6$ in the second group.
For training we balance the amount of positive samples (from the desired class) and negative samples formed from the remaining classes, i.e. $16$ and $5$ for two groups accordingly.
In the group of snakes, we use $160$ versus $160$ images randomly drawn from the desired class and from the remaining $16$ classes ($10$ from each). 
We used the same amount of samples for testing. 
In the group of balls, we use $100$ versus $100$ images for training randomly drawn from the desired class and from the remaining $5$ classes ($20$ from each). 
To keep the setting similar across datasets, we used double amount of samples for testing. 
To get better statistics of the performance we repeat each train/test split $20$ times.
\begin{table*}[thp]
\centering
\scalebox{1}{
\setlength{\tabcolsep}{4pt}
\begin{tabular}{c|l|c|c|c|c|c|c}		          				          
&			& {\tt SVM}  		& {\tt Margin Transfer} & {\tt SVM+}  		&& {\tt Reference}   	\\
&			&  image 		&  image+bbox	 	& image+bbox   		&& \it{(SVM bbox)}	 \\\hline
1 & Thunder snake 	& $65.76 \pm 0.85$ 	& $\bf{66.28 \pm 0.86}$ & $65.43 \pm 0.82$ 	&& $\it{66.51 \pm 0.85}$\\
2 & Ringneck snake 	& $67.28 \pm 0.67$ 	& $67.68 \pm 0.62$ 	& $\done\bf{68.40 \pm 0.60}$ && $\it{68.45 \pm 0.48}$\\
3 & Hognose snake 	& $65.10 \pm 0.69$ 	& $65.28 \pm 0.66$ 	& $\done\bf{68.32 \pm 0.43}$ && $\it{65.92 \pm 0.62}$\\
4 & Green snake 	& $73.42 \pm 0.44$ 	& $\done\bf{73.89 \pm 0.38}$ & $73.40 \pm 0.40$ 	&& $\it{74.46 \pm 0.36}$\\
5 & King snake 		& $76.31 \pm 0.45$ 	& $76.67 \pm 0.41$ 	& $\done\bf{78.51 \pm 0.48}$ && $\it{77.56 \pm 0.46}$\\
6 & Garter snake 	& $72.92 \pm 0.55$ 	& $73.20 \pm 0.47$ 	& $\done\bf{76.70 \pm 0.53}$ && $\it{75.35 \pm 0.52}$\\
7 & Water snake 	& $67.48 \pm 0.55$ 	& $67.75 \pm 0.57$ 	& $\done\bf{68.40 \pm 0.42}$ && $\it{65.35 \pm 0.52}$\\
8 & Vine snake 		& $79.42 \pm 0.33$ 	& $79.26 \pm 0.33$ 	& $\done\bf{79.98 \pm 0.29}$ && $\it{80.67 \pm 0.46}$\\
9 & Night snake 	& $57.42 \pm 0.65$ 	& $57.62 \pm 0.70$ 	& $\bf{58.07 \pm 0.57}$ && $\it{56.51 \pm 0.53}$\\
10 & Boa constrictor 	& $72.85 \pm 0.57$ 	& $72.68 \pm 0.53$ 	& $\done\bf{75.34 \pm 0.60}$ && $\it{72.62 \pm 0.49}$\\
11 & Rock python 	& $63.26 \pm 0.59$ 	& $63.29 \pm 0.64$ 	& $\done\bf{65.79 \pm 0.44}$ && $\it{63.20 \pm 0.41}$\\
12 & Indian cobra 	& $64.29 \pm 0.57$ 	& $\bf{64.59 \pm 0.59}$ & $64.51 \pm 0.62$ 	&& $\it{63.29 \pm 0.69}$\\
13 & Green mamba 	& $72.56 \pm 0.46$ 	& $72.89 \pm 0.49$ 	& $\done\bf{73.14 \pm 0.50}$ && $\it{73.60 \pm 0.51}$\\
14 & Sea snake 		& $80.29 \pm 0.41$ 	& $80.23 \pm 0.38$ 	& $\bf{80.82 \pm 0.37}$ && $\it{77.67 \pm 0.36}$\\
15 & Horned viper 	& $69.75 \pm 0.51$ 	& $69.73 \pm 0.48$ 	& $\done\bf{71.43 \pm 0.55}$ && $\it{72.53 \pm 0.42}$\\
16 & Diamondback 	& $75.39 \pm 0.50$ 	& $75.64 \pm 0.43$ 	& $\done\bf{77.21 \pm 0.45}$ && $\it{76.01 \pm 0.51}$\\
17 & Sidewinder 	& $\bf{68.85 \pm 0.42}$ & $68.53 \pm 0.57$ 	& $68.53 \pm 0.59$ 	&& $\it{69.84 \pm 0.57}$\\
\end{tabular} }
\centering
\parbox{13cm}{
 \caption{ImageNet dataset, group of snakes (bounding box annotation as privileged information). 
The numbers are mean and standard error of accuracy over $20$ runs. 
The best result is highlighted in \textbf{boldface}. Highlighted \highlight{blue} indicates significant improvement of the methods that utilize privileged information (Margin Transfer and/or SVM+) over the methods that do not (SVM). We used a paired Wilcoxon test with $95\%$ confidence level as a reference.
Additionally, we also provide the SVM performance on $\Xcal^{*}$ (last column). 
\label{tab:ImNet_snake}
}}
\end{table*}
\begin{table*}[thp]
\centering
\scalebox{1}{
\setlength{\tabcolsep}{4pt}
\begin{tabular}{c|l|c|c|c|c|c|c}		          				          
&			& {\tt SVM}  		& {\tt Margin Transfer} & {\tt SVM+}  		&& {\tt Reference}   	\\
&			&  image 		&  image+bbox	 	& image+bbox   		&& \it{(SVM bbox)}	 \\\hline
1 & Soccer ball 	& $65.95 \pm 0.66$ 	& $65.95 \pm 0.66$ 	& $\done\bf{67.42 \pm 0.67}$ && $\it{69.78 \pm 0.44}$\\
2 & Croquet ball 	& $73.31 \pm 0.38$ 	& $73.70 \pm 0.39$ 	& $\done\bf{73.80 \pm 0.40}$ && $\it{74.76 \pm 0.39}$\\
3 & Golf ball 		& $\bf{76.46 \pm 0.47}$ & $76.18 \pm 0.52$ 	& $75.95 \pm 0.52$ 	&& $\it{68.53 \pm 0.46}$\\
4 & Ping-pong ball 	& $71.80 \pm 0.54$ 	& $71.71 \pm 0.50$ 	& $\done\bf{72.85 \pm 0.44}$ && $\it{71.20 \pm 0.59}$\\
5 & Rugby ball 		& $76.08 \pm 0.40$ 	& $76.00 \pm 0.43$ 	& $\done\bf{82.90 \pm 0.29}$ && $\it{71.07 \pm 0.57}$\\
6 & Tennis ball 	& $67.57 \pm 0.48$ 	& $67.65 \pm 0.44$ 	& $\done\bf{68.17 \pm 0.45}$ && $\it{65.36 \pm 0.71}$\\
\end{tabular} }
\parbox{13cm}{
 \caption{ImageNet dataset, group of sport balls (bounding box annotation as privileged information). 
The numbers are mean and standard error of accuracy over $20$ runs. 
The best result is highlighted in \textbf{boldface}. Highlighted \highlight{blue} indicates significant improvement of the methods that utilize privileged information (Margin Transfer and/or SVM+) over the methods that do not (SVM). We used a paired Wilcoxon test with $95\%$ confidence level as a reference.
Additionally, we also provide the SVM performance on $\Xcal^{*}$ (last column).
\label{tab:ImNet_ball}
}}
\end{table*}

\textbf{Results.} 
As we can see from Table~\ref{tab:ImNet_snake} and Table~\ref{tab:ImNet_ball} utilizing bounding box annotation as privileged information for fine-grained classification is useful. 
In both tables, the LUPI methods outperform the non-LUPI SVM baseline in all but $1$ case. 
In the group of snakes, SVM+ clearly outperforms SVM in $14$ cases, and Margin Transfer outperforms SVM in $12$ cases out of $17$. 
In this experiment, the SVM+ method is able to exploit the privileged information much better than the Margin Transfer method (in $13$ out of $17$ cases, and $1$ tie in the case of Sidewinder snake). 
In the group of balls, we observe very similar results with clear advantage of the SVM+ method over all other methods. 
Margin Transform shows only minor difference with respect to standard SVM. 

Noticeably, the performance in the privileged space is not superior to the original data space, sometimes it is even worse, especially in the group of balls. 
Since our Margin transfer method relies directly on the performance in the privileged space, 
its ability to exploit easy and hard samples is limited in this scenario. 
On the other hand, modeling the slacks in the form of regression model, as SVM+ does, works well. 
We suspect it is more suitable when privileged and original spaces are of the same modality, i.e. here, the privileged information is obtained from a subset of the same image features that are used for the original data representation. 

\subsection{Textual description as privileged information}
\label{sec:text}
A textual description provides complementary view to a visual representation of an object. 
This can be used as privileged information in object classification task. 
We use two datasets to explore textual description as the source of privileged information and we will describe them in turn. 
The first dataset is \emph{IsraelImages}\footnote{\url{http://people.cs.umass.edu/~ronb/image_clustering.html}} introduced in \cite{Bek07}. 
The dataset has $11$ classes, $1823$ images in total, with a textual description (up to $18$ words) attached to each of the image. 
The number of samples per class is relatively small, around $150$ samples, and varies from $96$ to $191$ samples. 
We merge the classes into three groups: nature (birds, trees, flowers, desert), religion (christianity, islam, judaism, symbols) and urban (food, housing, personalities), 
and perform binary classification on the pairs of groups.  
We use $L_2$ normalized $4096$-dimensional Fisher vectors~\cite{PerSanMen10} extracted from the images as the original data representation and bag-of-words representation of the text data as privileged information. 
We use $100$ images per group for training and $200$ per group for testing.  
We repeat the train/test split $20$ times. 

The second dataset is a \emph{Accessories} dataset\footnote{\url{http://tamaraberg.com/attributesDataset/index.html}} introduced in \cite{BerBerShi10}. 
The dataset contains products taken from variety of e-commerce sources with the images and their textual descriptions. 
The products are grouped into $4$ broad shopping categories: \textit{bags, earrings, ties, and shoes}. 
We randomly select $1800$ samples from this dataset for our experiments, $450$ samples from each category. 
We generated $6$ binary classification tasks for each pair of the $4$ classes with $100$ samples per class for training and $200$ per class for testing. 
This second dataset contains longer text descriptions than the \emph{IsraelImages} dataset. 
These longer texts allow us to use advanced features introduced recently in term of word vectors instead of simple term frequency features.
We extracted $200$ dimensional word vector using a neural network skip-gram architecture \cite{MikSutCheCoretal13}\footnote{\url{https://code.google.com/p/word2vec/}}. Then we constructed a codebook of $100$ word-vector to convert this word representation into a fixed-length sentence representation and apply $L_1$ normalization. 
We use $L_1$ normalized SURF descriptors \cite{Bay08} extracted from images with $100$ visual words codebook as feature representation in the original space.
\begin{table*}[thp!]
\centering
\scalebox{1}{
\setlength{\tabcolsep}{4pt}
\begin{tabular}{c|l|c|c|c|c|c|c|c|c|c}				          				          
&				& {\tt SVM}  		& {\tt Margin Transfer} & {\tt SVM+}  		&& {\tt Reference}   	\\
&				&  image 		&  image+text	        & image+text   		&& \it{(SVM text)}	\\\hline
1 & Nature versus Religion 	& $81.01 \pm 0.49$ 	& $\bf{81.15 \pm 0.51}$ & $81.08 \pm 0.50$ 	&& $\it{84.52 \pm 0.47}$\\
2 & Religion versus Urban 	& $65.55 \pm 0.43$ 	& $\bf{65.62 \pm 0.50}$ & $65.52 \pm 0.57$ 	&& $\it{88.12 \pm 0.44}$\\
3 & Nature versus Urban 	& $78.97 \pm 0.42$ 	& $78.73 \pm 0.49$ 	& $\bf{79.15 \pm 0.53}$ && $\it{83.11 \pm 0.56}$\\
\end{tabular} }
\caption{Israeli dataset (textual description as privileged information). The numbers are mean and standard error of the accuracy over $20$ runs. 
As reference we also provide the SVM performance on the $\Xcal^{*}$ (last column).}
\label{tab:Israeli}
\end{table*}
\begin{table*}[thp!]
\centering
\scalebox{1}{
\setlength{\tabcolsep}{4pt}
\begin{tabular}{c|l|c|c|c|c|c|c|c|c|c}				          				          
&				& {\tt SVM}  		& {\tt Margin Transfer} & {\tt SVM+}  		&& {\tt Reference}   	\\
&				&  image 		&  image+text	        & image+text   		&& \it{(SVM text)}	\\\hline
1 & Earrings versus Bags 	& $90.25 \pm 0.27$ 	& $\bf{90.48 \pm 0.30}$ & $89.73 \pm 0.32$ 	&& $\it{97.87 \pm 0.18}$\\
2 & Earrings versus Shoes 	& $92.65 \pm 0.20$ 	& $\bf{92.81 \pm 0.13}$ & $92.46 \pm 0.23$ 	&& $\it{98.58 \pm 0.16}$\\
3 & Bags versus Ties 		& $87.70 \pm 0.44$ 	& $\bf{88.08 \pm 0.43}$ & $87.80 \pm 0.45$ 	&& $\it{98.62 \pm 0.13}$\\
4 & Bags versus Shoes 		& $\bf{90.52 \pm 0.24}$ & $90.28 \pm 0.25$ 	& $88.61 \pm 0.40$ 	&& $\it{96.26 \pm 0.25}$\\
5 & Ties versus Earrings 	& $\bf{90.27 \pm 0.28}$ & $90.11 \pm 0.24$ 	& $88.97 \pm 0.25$ 	&& $\it{99.57 \pm 0.07}$\\
6 & Ties versus Shoes 		& $\bf{81.60 \pm 0.61}$ & $81.53 \pm 0.48$ 	& $80.90 \pm 0.51$ 	&& $\it{98.96 \pm 0.15}$\\
\end{tabular} }
\caption{Accessories dataset (textual description as privileged information). 
The numbers are mean and standard error of the accuracy over $20$ runs. 
As reference we also provide the SVM performance on the $\Xcal^{*}$ (last column).}
\label{tab:Clothes}
\end{table*}

\textbf{Results.} 
As we can see from Table \ref{tab:Israeli} and Table \ref{tab:Clothes} utilizing textual privileged information as provided in the \emph{IsraelImages} and \emph{Accessories} datasets does not help. 
All methods, LUPI and non-LUPI have near equal performance, and there is no signal of privileged information being utilized in both LUPI methods. 
This might seem contradictory to the high performance of the reference baseline in the text domain, $\Xcal^{*}$.
However high accuracy in the privileged space does not necessarily mean that the privileged information is helpful. 
For example, assume we used the labels themselves as privileged modality: classification would be trivial, but it would provide no additional information to transfer.
In the \emph{IsraelImages}, the textual descriptions of the images are very sparse and contain many duplicates, and in the \emph{Accessories} datasets the texts are ``too easy''. 
Therefore, the margin distance in the privileged space does not capture the easiness and hardness of different samples, and mainly preserves the class separation only. The performance does not degrade nevertheless. 

\section{Multiclass classification}
\label{sec:multiclass}
We also explore the benefits of utilizing the LUPI methods in the multiclass setup with one-versus-rest learning strategy. 
We train one binary classifier for each class to distinguish samples of this class (positive label) versus samples from the remaining classes (negative label). 
For a test point, the label is assigned based on the class with maximum prediction value over all binary classifiers. 
For model selection, we use 5 fold cross-validation scheme and search over range of regularization parameters the same as before. 
In order to calibrate the prediction scores from different classifiers we use one parameter value to train all the binary classifiers, and cross validate the multiclass performance. 
The best parameter (pair of parameters) is used to retrain all classifiers. 
To the best of our knowledge, this is the first time LUPI methods are studied in the classification learning setting with more than two classes.
We run the multiclass setting on all the datasets described previously for the binary classification task. 
The results are summarized in Table~\ref{tab:multi}.
\begin{table*}[thp]
\centering
\scalebox{0.85}{
\setlength{\tabcolsep}{4pt}
\begin{tabular}{c|l|l|c|c|c|c|c|c|c|c|c}				          				          
&{\tt Dataset}			& {\tt privileged} 	& {\tt SVM}  	& {\tt Margin Transfer} & {\tt SVM+}		&& {\tt Reference}   	\\\hline
& AwA ($10$ classes)		& attributes 	& $45.41 \pm 0.18$ 	& $\bf{46.44 \pm 0.27}$ & $42.07 \pm 0.36$	&& $\it{56.18 \pm 0.21}$\\
& Snakes ($17$ classes)		& bbox		& $30.41 \pm 0.18$ 	& $\bf{31.61 \pm 0.19}$ & $31.09 \pm 0.24$ 	&& $\it{31.84 \pm 0.15}$\\
& Sport Balls ($6$ classes)	& bbox		& $51.78 \pm 0.26$ 	& $51.65 \pm 0.36$ 	& $\bf{52.75 \pm 0.35}$	&& $\it{49.47 \pm 0.29}$\\
& Israeli ($3$ groups)		& text		& $60.16 \pm 0.41$	& $\bf{60.65 \pm 0.46}$	& $60.14 \pm 0.42$ 	&& $\it{76.37 \pm 0.43}$\\
& Accessories ($4$ classes)	& text		& $76.45 \pm 0.28$	& $\bf{76.48 \pm 0.26}$	& $72.68 \pm 0.37$	&& $\it{97.00 \pm 0.16}$\\
\end{tabular} }
 \caption{Multiclass performance. The numbers are mean and standard error of accuracy over $20$ runs. 
The best result is highlighted in \textbf{boldface}. 
Additionally, as reference we also provide the performance on $\Xcal^{*}$ (last column). 
}
\label{tab:multi}
\end{table*}

\textbf{Results.} 
As we can see from Table~\ref{tab:multi} utilizing privileged information is useful for multiclass classification. 
The LUPI methods outperform the non-LUPI baseline (SVM) in all datasets. 
Overall the Margin Transfer is superior to SVM+ in all but one case (Sport Balls); it is more stable contrary to performance drop of the SVM+ in the \emph{AwA} and \emph{Accessories} datasets; and it follows the tendency to outperform SVM when performance in the privileged space (column Reference) is better than in the original space (column SVM).

\section{Human annotation as privileged information}
\label{sec:human}
For this experiment, we collect Mechanical Turk\footnote{\scriptsize{\url{https://www.mturk.com/mturk/}}} annotation of images to define easy and hard samples for human learning. 
We analyze the advantages of having this information in comparison to the LUPI methods. 
We managed to collect reliable human annotation for $8$ out of $10$ classes in the AwA dataset: \emph{chimpanzee, giant panda, leopard, persian cat, hippopotamus, raccoon, rat, seal} except the class \emph{pig} and \emph{humpback whale}.
In the dataset, many pictures of the class \emph{pig} are related to food product rather than animal itself, which create some aesthetic issues for user to make an objective judgment. Images in the class \emph{humpback whale} lack variability across samples, which makes it difficult to distinguish between easy and hard ones. So we make the analysis based on the annotation from $8$ classes. 
In our user study, the participant is shown a set of images of one particular class and is asked to select the most prominent (easiest) images first, then proceed to less obvious, and so on, until most difficult samples are left. We aggregate these ranking information across overlapping sets of images to compute a global order of images per category. The score is in the range from $1$ (hardest) till $16$ (easiest). 
We observe, that most of the time the easiest instances are those with clearly visible object of interest in the center of the image, where as the hardest are occluded objects, small sized, or with humans. 

\textbf{Evaluation.}
First, we analyze the advantage of transferring the information from human annotation by looking into accuracy performance. 
In order to map the easy-hard score into the margin distance $\rho_i$, we use linear scaling of the score to $[0, 2]$ interval, so values between $0$ and $1$ correspond to hard samples and values greater than $1$ correspond to easy samples. 
After this we proceed directly to the second stage of the Margin Transfer method \eq{eq:margintransfer}. 
We report the results over $28$ pairs of classes in the Table~\ref{tab:AwA_human} (last column). 

\begin{figure*}[thp!!!]
  \includegraphics[width=0.99\columnwidth]{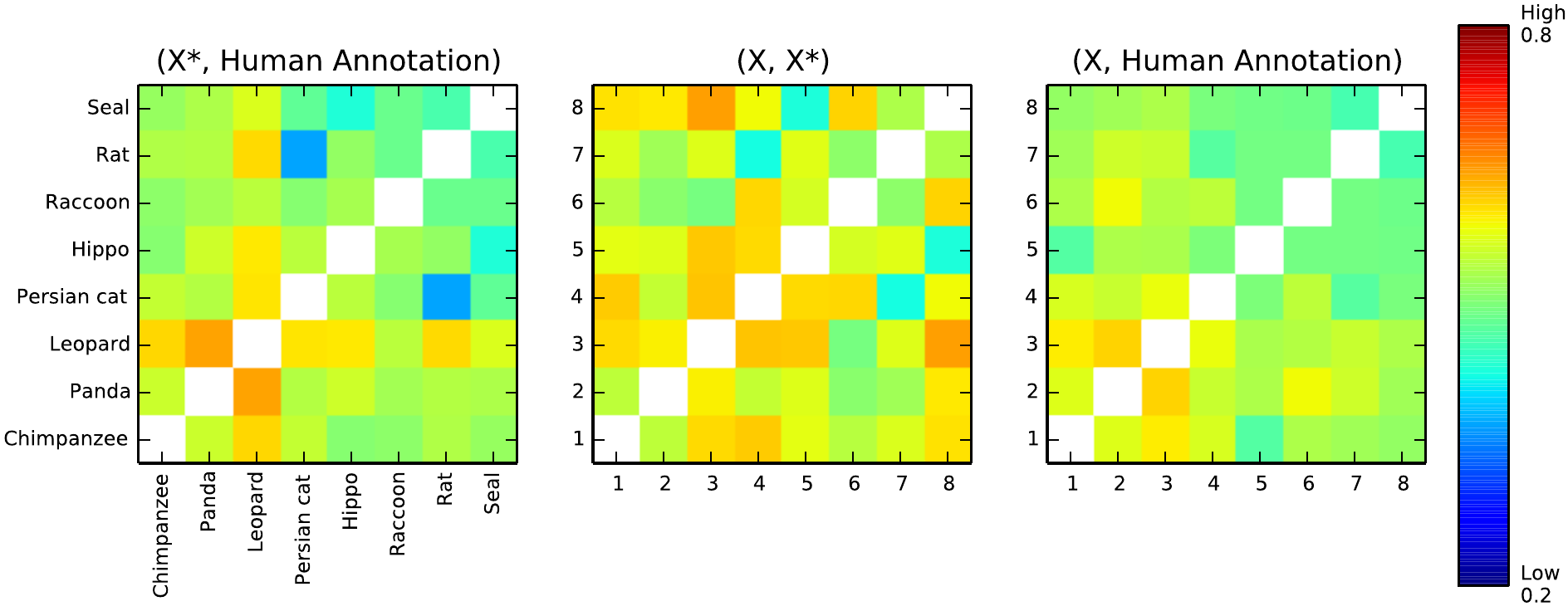}
\caption{Human annotation as privileged information. Kendall tau rank correlation analysis is used to explore the correlation between easy-hard samples defined by human annotation and easy and hard samples that we identify when learning in the privileged space (left). 
We analyze the correlation between easy and hard samples in the privileged and original data spaces (center), and between human annotation and samples in the original data space (right).
} 
\label{fig:AwA_human}
\end{figure*}

Secondly, we study whether the easy-hard score in the human understanding correlates with easy and hard samples that we identify when learning in the privileged space of attributes.
We use the Kendall tau rank correlation analysis and compute the correlation coefficient across $28$ learning tasks. 
For each task, we compute the correlation between the margin distance from the training samples to the classifying hyperplane in the privileged space of attributes, $\inner{\w^\ast}{\x_i^\ast}$, and the easy-hard scores obtained from the human annotation $\textbf{score}_i \: y_i$. 
%
Similarly we evaluate the correlation between easy and hard samples in the original and privileged spaces that the Margin Transfer method relies on.  For this, we compute the correlation between the predicted values of the classifiers trained on data $X$ from the original space and on $X^\ast$ from the privileged space of attributes. To complete this analysis we also look at correlation between user defined easy-hard scores and easy and hard samples in the original space. 
For visualization, we aggregate the results into a symmetric table, where each entry is the tau coefficient computed for the corresponding pair of classes in the binary learning task, refer to Figure~\ref{fig:AwA_human}. 

\begin{table*}[thp!!!]
\centering
\scalebox{0.85}{
\setlength{\tabcolsep}{4pt}
\begin{tabular}{c|l|c|c|c|c|c|c|c}				          				          
&					& {\tt SVM}  		& {\tt Margin Transfer}  & {\tt Margin Transfer}  \\
&					&  image 		&  image+attributes	 & image+human annot.  	  \\\hline
1 & Chimpanzee versus Giant panda 	& $83.25 \pm 0.53$ 	& $83.40 \pm 0.43$ 	& $\bf{83.72 \pm 0.58}$ 	\\
2 & Chimpanzee versus Leopard 		& $86.63 \pm 0.35$ 	& $\bf{86.71 \pm 0.38}$ & $86.43 \pm 0.30$ 	\\
3 & Chimpanzee versus Persian cat 	& $83.91 \pm 0.46$ 	& $\bf{84.22 \pm 0.41}$ & $83.93 \pm 0.38$ 	\\
4 & Chimpanzee versus Hippopotamus 	& $81.05 \pm 0.28$ 	& $\bf{81.90 \pm 0.27}$ & $80.88 \pm 0.29$ 	\\
5 & Chimpanzee versus Raccoon 		& $80.11 \pm 0.50$ 	& $\bf{81.17 \pm 0.48}$ & $80.76 \pm 0.55$ 	\\
6 & Chimpanzee versus Rat 		& $80.15 \pm 0.43$ 	& $\bf{81.22 \pm 0.43}$ & $79.91 \pm 0.42$ 	\\
7 & Chimpanzee versus Seal 		& $\bf{85.80 \pm 0.26}$ & $85.43 \pm 0.44$ 	& $85.60 \pm 0.38$ 	\\
8 & Giant panda versus Leopard 		& $87.82 \pm 0.32$ 	& $87.32 \pm 0.37$ 	& $\bf{88.11 \pm 0.36}$ 	\\
9 & Giant panda versus Persian cat 	& $87.66 \pm 0.37$ 	& $86.35 \pm 0.30$ 	& $\bf{88.12 \pm 0.28}$ 	\\
10 & Giant panda versus Hippopotamus 	& $85.36 \pm 0.41$ 	& $84.96 \pm 0.51$ 	& $\bf{85.90 \pm 0.45}$ 	\\
11 & Giant panda versus Raccoon 	& $83.52 \pm 0.44$ 	& $83.17 \pm 0.44$ 	& $\bf{83.77 \pm 0.52}$ 	\\
12 & Giant panda versus Rat 		& $81.76 \pm 0.45$ 	& $81.76 \pm 0.38$ 	& $\bf{82.20 \pm 0.44}$ 	\\
13 & Giant panda versus Seal 		& $85.47 \pm 0.41$ 	& $85.58 \pm 0.40$ 	& $\bf{85.72 \pm 0.37}$ 	\\
14 & Leopard versus Persian cat 	& $\bf{90.18 \pm 0.23}$ & $89.87 \pm 0.26$ 	& $89.76 \pm 0.31$ 	\\
15 & Leopard versus Hippopotamus 	& $86.37 \pm 0.33$ 	& $86.10 \pm 0.31$ 	& $\bf{86.43 \pm 0.35}$ 	\\
16 & Leopard versus Raccoon 		& $77.40 \pm 0.51$ 	& $76.53 \pm 0.76$ 	& $\bf{78.21 \pm 0.47}$ 	\\
17 & Leopard versus Rat 		& $81.82 \pm 0.26$ 	& $81.85 \pm 0.40$ 	& $\bf{82.11 \pm 0.33}$ 	\\
18 & Leopard versus Seal 		& $87.28 \pm 0.36$ 	& $\bf{87.62 \pm 0.36}$ & $87.56 \pm 0.36$ 	\\
19 & Persian cat versus Hippopotamus 	& $84.85 \pm 0.56$ 	& $85.13 \pm 0.46$ 	& $\bf{85.30 \pm 0.48}$ 	\\
20 & Persian cat versus Raccoon 	& $\bf{84.50 \pm 0.44}$ & $84.33 \pm 0.44$ 	& $84.42 \pm 0.51$ 	\\
21 & Persian cat versus Rat 		& $65.32 \pm 0.38$ 	& $\bf{66.36 \pm 0.51}$ & $65.05 \pm 0.43$ 	\\
22 & Persian cat versus Seal 		& $\bf{80.61 \pm 0.38}$ & $79.77 \pm 0.48$ 	& $79.80 \pm 0.44$ 	\\
23 & Hippopotamus versus Raccoon 	& $80.61 \pm 0.61$ 	& $\bf{81.88 \pm 0.60}$ & $81.20 \pm 0.62$ 	\\
24 & Hippopotamus versus Rat 		& $77.83 \pm 0.45$ 	& $\bf{79.70 \pm 0.41}$ & $78.07 \pm 0.39$ 	\\
25 & Hippopotamus versus Seal 		& $67.91 \pm 0.60$ 	& $\bf{68.62 \pm 0.55}$ & $66.96 \pm 0.61$ 	\\
26 & Raccoon versus Rat 		& $\bf{73.75 \pm 0.38}$ & $73.66 \pm 0.43$ 	& $73.60 \pm 0.35$ 	\\
27 & Raccoon versus Seal 		& $83.86 \pm 0.42$ 	& $\bf{84.53 \pm 0.41}$ & $84.06 \pm 0.41$ 	\\
28 & Rat versus Seal 			& $74.03 \pm 0.56$ 	& $\bf{74.83 \pm 0.52}$ & $74.31 \pm 0.47$ 	
\end{tabular} }
\caption{Human annotation as privileged information. 
We incorporate human perception of easiness and hardness into the margin distance and perform Margin Transfer with human annotation (last column). The numbers are mean accuracy and standard error over $20$ runs.
}
\label{tab:AwA_human}
\end{table*}

\textbf{Results.} 
As we can see from Table~\ref{tab:AwA_human}, collecting good quality human annotation can help to improve the classification performance, however it cannot solve the problem of negative transfer from privileged space to the original space. 
In some cases it clearly helps, as to classify the categories \emph{Giant panda} and \emph{Leopard} versus others, 
and in other cases it does not, as in the category \emph{Chimpanzee}. 

As we can see from Figure~\ref{fig:AwA_human}, overall the correlation between easy-hard samples in the privileged and the original data spaces (center) has higher signal than between human annotation versus privileged data space (left) and versus original data space (right). 
If we look closely at the case with \emph{Giant panda} (second column across the tables), we observe that indeed the correlation between human annotation and ranking in the original data $X$ (right) is more expressed than between $X^*$ and $X$ (center), which possibly explains the performance gain when using human annotation as privileged information instead of attribute description.
It is not always the case for the \emph{Leopard} class, when for example, in classification versus \emph{Seal} the correlation ($X$, $X^*$) is considerably stronger than ($X$, Human annotation), which also matches better performance gain when utilizing attributes as privileged information. The class \emph{Chimpanzee} (first column across the tables) seems to be more suitable to explore the privileged information based on attribute description (center) comparing to human annotation (right). 
As we can see, in other classes on the right plot, there is little signal in the correlation ($X$, Human annotation), i.e. mostly blue color, which coincides with rather disadvantageous performance when doing Margin Transfer with human annotation as privileged information in comparison to Margin Transfer with attributes as privileged information. 
Low correlation (striking blue color), like in pairs \emph{Rat versus Persian cat}, also \emph{Seal versus Hippopotamus}, can be explained by low performance of the classifiers on this pairs. Low performance in our case means a lot of hard/misclassified samples that are on the wrong side of the classifier hyperplane which influences the margin score. In principle, this situation is not suitable for our ranking correlation analysis, because human defined easy hard sample scores do not account for such misclassifications.

\section{Conclusion and Future Work}
\label{sec:conclusion}
We have studied the setting of learning using privileged information (LUPI) in visual object classification tasks. 
We showed how it can be applied to several situations that previously were handled by hand-crafted separate methods. 
Our experiments show that prediction performance often improves when utilizing the privileged information in both, binary and multiclass learning settings. 
We have studied two approaches for solving the LUPI task: SVM+ and the proposed Margin Transfer method. 
Margin Transfer shows comparable performance to the SVM+ algorithm and can be easily applied using standard SVM solvers. 
The analysis of main modeling assumptions of both LUPI methods suggested that privileged information can be utilized 
as measure of easiness and hardness of the samples, and this can guide learning of the classifier to better generalization. 
Also in this work, we made an attempt to understand what is the essence of easy and hard samples in the learning process, and we compared them with human labeled easy and hard samples annotated during user studies. 

In the future, we are interested in exploring the relatedness of multiple tasks \cite{quadrianto2010multitask,tommasi2013beyond} in privileged space and original space.
Another direction is to explore the possibility to predict the priviledged data at test time. 
%

\paragraph{Acknowledgments}
This work was in parts funded by the European Research Council under the European Union's 
Seventh Framework Programme (FP7/2007-2013)/ERC grant agreement no 308036.


\bibliographystyle{unsrt}
\bibliography{bibfile}
\end{document}